\pdfoutput=1

\documentclass[11pt]{article}

\usepackage[]{emnlp2021}

\usepackage{times}
\usepackage{latexsym}

\usepackage{graphicx}
\usepackage{textcomp}
\usepackage{xcolor}
\usepackage{tabularx, booktabs}
\usepackage{multirow}
\usepackage{siunitx}
\usepackage{subcaption}
\usepackage{placeins}
\usepackage{float}
\usepackage[export]{adjustbox}
\usepackage{comment}
\usepackage[labelfont=bf]{caption}

\usepackage[T1]{fontenc}

\usepackage[utf8]{inputenc}

\usepackage{microtype}

\title{Towards Human-Free Automatic Quality Evaluation of German Summarization}

\author{Neslihan Iskender \\
  Technische Universität Berlin \\ 
  Quality and Usability Lab\\ 
  \texttt{neslihan.iskender}\\\texttt{@tu-berlin.de} \\\And
  Oleg Vasilyev \\
  Primer Technologies Inc. \\
  San Francisco, California \\
  \texttt{oleg@primer.ai} \\ \And
  Tim Polzehl \\
  Technische Universität Berlin \\ 
  Quality and Usability Lab\\ 
  \texttt{tim.polzehl1}\\\texttt{@tu-berlin.de} \\
  \AND
  John Bohannon \\
  Primer Technologies Inc. \\
  San Francisco, California \\
  \texttt{john@primer.ai} \\ \And
  Sebastian Möller \\
  Technische Universität Berlin \\ 
  Quality and Usability Lab\\ 
  \texttt{sebastian.moeller@tu-berlin.de}}

\begin{document}

\maketitle

\begin{abstract}
Evaluating large summarization corpora using humans has proven to be expensive from both the organizational and the financial perspective. Therefore, many automatic evaluation metrics have been developed to measure the summarization quality in a fast and reproducible way. However, most of the metrics still rely on humans and need gold standard summaries generated by linguistic experts. Since BLANC does not require golden summaries and supposedly can use any underlying language model, we consider its application to the evaluation of summarization in German. This work demonstrates how to adjust the BLANC metric to a language other than English. We compare BLANC scores with the crowd and expert ratings, as well as with commonly used automatic metrics on a German summarization data set. Our results show that BLANC in German is especially good in evaluating informativeness.
\end{abstract}

\section{Introduction}

\label{sec:intro}

In text summarization, the human evaluation conducted by linguistic experts is the gold standard, but it is expensive and slow \cite{bhandari2020}. The wide range of quality evaluation criteria and non-standard annotation set-ups used by researchers lead to conflicting and mostly non-reproducible results \cite{howcroft2020}. Therefore, researchers have focused on developing automatic summarization evaluation metrics relying on gold standard summaries written by humans, such as BLEU \cite{papineni2002bleu}, ROUGE \cite{lin2004rouge} and BERTScore \cite{zhang2020}, or human-free quality metrics such as Jensen-Shannon (JS) \cite{louis2009}, SUM-QE \cite{xenouleas2019} and  BLANC \cite{vasilyev2020}.

Currently, ROUGE is still the most popular automatic summarization quality evaluation metric in ACL conferences \cite{gao2019}. Its requirement of gold standard summaries is limiting researchers to use only the available data sets with gold standard summaries such as the TAC and CNN/Daily Mail data sets derived from high-quality English texts in the news domain \cite{dang2009, nallapati2016}. 
To overcome this limitation, we focus in this paper on the human-free summarization quality evaluation metric BLANC, which so far has been applied only to a limited number of English data sets. We demonstrate how to adjust BLANC to a language other than English on the example of the German language, and we compare its performance to the commonly used automatic metrics and human annotations.

\section{Related Work}
\label{sec:related-work}

The summarization quality evaluation is crucial for the development of automatic summarization systems to determine whether the system has properly outperformed the existing tools in terms of quality and speed or whether the designed properties work as intended \cite{van2018}. Traditional summarization evaluation methods can be categorized into intrinsic evaluation and extrinsic evaluation \cite{belz2006comparing}. 

In intrinsic evaluation, the summarization quality evaluation is directly based on the summary itself without considering the source text, typically carried out as pair comparison (compared to expert summaries) or using absolute scales without having a reference \cite{jones1995}. However, the extrinsic evaluation is based on measuring the summary's impact on the completion of some task using the source document \cite{mani2001b}. Both evaluation methods can be performed by humans, usually linguistic experts, as well as by automatic metrics. 

Despite the high cost, subjectivity and lack of standards, the human evaluation is still considered to be the gold standard in summarization evaluation \cite{howcroft2020, bhandari2020}. The lack of standards results in usage of different quality criteria, experimental designs and reporting schema across the research papers published in ACL and INLG conferences \cite{van2019best}. Therefore, automatic evaluation metrics are being introduced, such as  BLEU \cite{papineni2002bleu}, METEOR \cite{denkowski2014}, ROUGE \cite{lin2004rouge}, ROUGE-WE \cite{ng2015better}, SUM-QE \cite{xenouleas2019}, APES \cite{Matan2019APES}, Summa-QA \cite{Thomas2019SummaQA} and FEQA \cite{Esin2020FEQA}, \cite{peyrard2017a}, BERTScore \cite{zhang2020}, MoverScore \cite{zhao2019} and SUPERT \cite{Yang2020SUPERT} and BLANC \cite{vasilyev2020}, while ROUGE is still being by far the most popular metric \cite{gao2019}.

The automated metrics have been criticized for inconsistent correlations with human evaluations, and for the limited research on few existing English-language data sets \cite{reiter2009investigation, graham2015, novikova2017we, peyrard2017b, gabriel2020}.
In this paper, we explore moving BLANC measure from English to German, as a measure that requires replacement of the underlying model. We compare it with human annotations and with several lexical overlap based automated metrics, which are easy to move to another language.   

\section{Methodology}
\label{sec:methodology}

\subsection{Human Evaluation}

For our data analysis, we used a German summarization data set with 50 summaries, crowd and expert evaluations for these 50 summaries. 

The corpus contains queries with an average word count of 7.78, the shortest one with four words, and the longest with 17 words; posts from a customer forum of Deutsche Telekom with an average word count of 555, the shortest one with 155 words, and the longest with 1005 words; and corresponding query-based extractive summaries with an average word count of 63.32, the shortest one with 24 words, and the longest one with 147 words. 

The human annotations of the data set contain annotations both from two linguistic experts and 24 different crowd workers per summary for the following factors on a 5-point Mean Opinion Score (MOS) scale: overall quality, intrinsic quality factors (grammaticality, non-redundancy, referential clarity, focus, structure \& coherence), and extrinsic quality factors (summary usefulness, post usefulness, and summary informativeness). Human annotators evaluated the first six quality factors intrinsically, meaning that they did not read the source document and only see the summary itself. During the evaluation of extrinsic quality factors, human annotators read the query, the source, and the summary. In this paper, we only focus on the extrinsic quality factors since they reflect the system’s performance on the task, so they represent the content quality of a summary and are more suitable for comparing human evaluation with content-based automatic quality metrics.

In this data set, the first extrinsic metric \textit{summary usefulness}, also called content responsiveness, is defined as the version introduced in DUC 2003 \cite{Nist2003} which investigates the summary's usefulness concerning how useful the extracted summary is to satisfy the given goal \cite{shapira2019}. Following, the second extrinsic metric \textit{post usefulness}, also called relevance assessment, examines if the source document contains relevant information about the given goal  \cite{mani2001b,conroy2008}. Lastly, the third extrinsic metric \textit{summary informativeness} answers the question of how much information from the source document is preserved in the extracted summary \cite{mani2001b}.

\subsection{Automatic Metrics}

We calculated the BLEU and ROUGE scores using the sumeval library\footnote{\url{https://github.com/chakki-works/sumeval}} for German and BERTScore using the bert-score library\footnote{\url{https://github.com/Tiiiger/bert\textunderscore score}} using bert-base-german-cased as the language model \cite{zhang2020}. All of these three metrics are reference-based, so two linguistic experts created the gold standard summaries, which have an average word count of 58.18, the shortest one with 14 words, and the longest with 112 words. Additionally, we calculated the reference-free Jensen-Shannon (JS) similarity as described in \citet{louis2009}.

\subsection{BLANC for German}

Moving BLANC from one language to another is simple: all that has to be done is a replacement of the underlying language model, trained for generic Cloze task. We explore using three German-language trained models: bert-base-german-cased\footnote{https://huggingface.co/bert-base-german-cased}, bert-base-german-dbmdz-cased and bert-base-german-dbmdz-uncased\footnote{https://github.com/dbmdz/berts}. The algorithm\footnote{https://github.com/PrimerAI/blanc} does not need to be changed. BLANC does depend weakly \cite{vasilyev2020} on several parameters defining the frequency of masking for the Cloze task.

The BLANC family is parametrized by a step between masked tokens, $gap$, and by a minimal length $L$ at which a token is allowed to be masked. For example, $gap=3$ would mean that each third token would be masked. The minimal length may differ for three different kinds of tokens, accordingly to the tokenization by the underlying language model: it is $L_{normal}$ for a whole-word token; it is $L_{lead}$ for the first token when a word is split into two or more tokens; and it is $L_{follow}$ for the follow-up tokens of the split word.

If correlations with human scores can be trusted, one can specify the parameters that provide higher correlations. The value $gap=6$ is special because it approximately corresponds to the frequency of masking in standard BERT training. The value $gap=2$ was found to give the best correlations with human scores in English \cite{Oleg2020Sensitivity}. We certainly can take the same parameters as a good configuration and be done. However, German words are on average longer than English words, and we can anticipate a possibility of slightly more beneficial thresholds on the length of the tokens to be masked. In order to have firm conclusions, we have explored around 100, and narrowed down to 72 promising configurations defined by the choice from the 3 models x 24 choices of parameters: $gap=2,6$; $L_{normal}=4,5,6$; $L_{lead}=1,2$; $L_{follow}=1,100$. The latter $L_{follow}$ choices mean either allowing masking all tokens or not masking any. 

\section{Results}
\label{sec:results}

Results are presented for the MOS values of the three extrinsic quality scores (summary usefulness (SU), post usefulness (PU), and summary informativeness (SI)) assessed by crowd workers and experts, as well as for automatic metrics BLEU, ROUGE-1, ROUGE-2, ROUGE-L, BERTScore (we use F-scores), Jensen-Shannon similarity, and different configurations of BLANC score.

\subsection{Comparing Human Evaluation with BLANC}

To test the normality of the human annotations and automatic metrics, we carried out Anderson Darling tests showing that the summary usefulness measured by the crowd, summary informativeness measured by experts and BLANC were not normally distributed (\textit{p} $<$ 0.05). Therefore, we apply Spearman rank-order correlation coefficients to compare human ratings with automatic metrics.

For our analysis, we correlated the summary usefulness, post usefulness and summary informativeness measured by the crowd and the experts with different BLANC configurations. For the correlations between crowd and BLANC configurations, we observed that BLANC configuration with \textit{language model} $=$ \textit{bert-base-german-dbmdz-cased}, $gap = 2$, $L_{normal} = 4$, $L_{lead} = 2$, $L_{follow} = 1$ has the highest significant correlation with summary usefulness ($\rho$ $=$ 0.350, p $<$ 0.05), post usefulness ($\rho$ $=$ 0.378, p $<$ 0.05), and summary informativeness ($\rho$ $=$ 0.413, p $<$ 0.05). 

When we look at the correlation betwen expert and BLANC configurations, we see that the same BLANC configuration has the highest significant correlation with summary usefulness ($\rho$ $=$ 0.349, p $<$ 0.05) and the second highest correlation with summary informativeness ($\rho$ $=$ 0.460, p $<$ 0.05). Blocking masking of follow-up tokens $L_{follow}=100$ increases the correlation with summary informativeness  ($\rho$ $=$ 0.485, p $<$ 0.05). There is no significant correlation between post usefulness measured by the experts and BLANC configurations. 
These results show that the single BLANC configuration is robust across different correlations with human annotations in German.

We also found that the top 5 configurations best correlated with summary informativeness with experts and the top 5 with the crowd are all employing the model \textit{bert-base-german-dbmdz-cased} with $gap = 2$, and differ only by the $L$ thresholds (which have low influence).
We recommend to use the language model \textit{bert-base-german-dbmdz-cased}, $gap = 2$, $L_{normal} = 4$, $L_{lead} = 2$, $L_{follow} = 1$ when calculating BLANC for German language. Also, these results are perfectly in line with the findings of \citet{Oleg2020Sensitivity} for the correlations of BLANC ($gap = 2$) with summary informativeness on the English data set. BLANC correlates significantly with summary informativeness for both languages indicating that BLANC reflect the summary informativeness in a language agnostic way. 

Further, to analyze the effect of the source document's length, summary length, and compression level (summary length divided by source length), we divide our data into two groups for each criterion by its mean. We only observe significant correlations by grouping mean summary length (M $=$ 63.32) for the correlation of summary informativeness measured by the expert with BLANC. In the short summary group (N $=$ 30), the correlation ($\rho$ $=$ 0.476, p $<$ 0.05) was lower than in the long summary group ($\rho$ $=$ 0.516, p $<$ 0.05) showing that BLANC performs better when the summary is longer.

\subsection{Comparing BLANC with Other Automatic Metrics}

For comparing BLANC with commonly used automatic metrics, we calculated the Spearman rank-order correlation coefficients of the summary informativeness measured by the expert with BLEU, ROUGE-1, ROUGE-2, ROUGE-L, BERTScore-F, JS similarity and top three configurations of BLANC selected based on the correlation with summary informativeness. As noted in the previous section, all BLANC top models turned out to use \textit{bert-base-german-dbmdz-cased} with $gap = 2$, hence our notations for BLANC $B\_$ in the figures include only $L$-thresholds, for example the name '$B\_L4\_Ll2\_Lf1$' means $L_{normal} = 4$, $L_{lead} = 2$ and $L_{follow} = 1$.

\begin{figure}[h!]
\includegraphics[width=1\columnwidth]{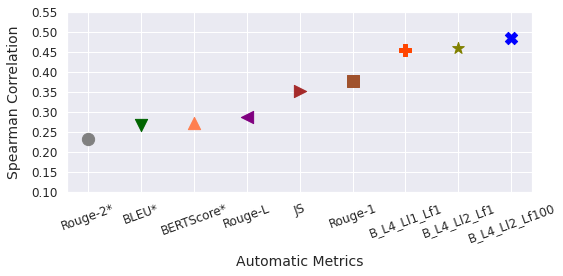}
{\footnotesize * not significant correlation}
\caption{Spearman Rank-order Correlation between Summary Informativeness measured by Experts and Automatic Metrics}
\label{fig:1}
\vspace{-0.5em}
\end{figure}

\begin{figure}[t!]
\includegraphics[width=1\columnwidth]{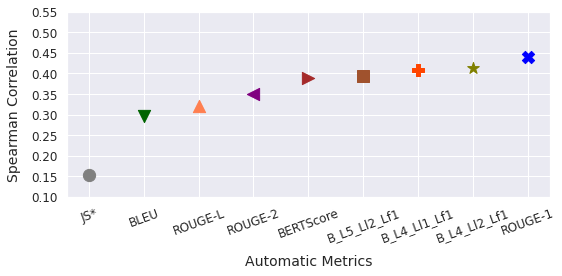}
{\footnotesize * not significant correlation}
\caption{Spearman Rank-order Correlation between Summary Informativeness measured by Crowd and Automatic Metrics}
\label{fig:2}
\vspace{-0.5em}
\end{figure}

Figure \ref{fig:1} shows the correlations of summary informativeness measured by the expert with automatic metrics. We observe that ROUGE-2, BLEU, and BERTScore do not correlate significantly with summary informativeness and the BLANC configuration with  \textit{language model} $=$ \textit{bert-base-german-dbmdz-cased}, $gap = 2$, $L_{normal} = 4$, $L_{lead} = 2$, $L_{follow} = 100$ outperforms all the automatic metrics. However, Figure \ref{fig:2} shows that Rouge-1 correlates better than BLANC with summary informativeness measured by the crowd.

Why the BLANC versions are much better with the experts, while ROUGE-1 is better with the crowd workers? We speculate that ROUGE-1 is a lexical overlap measure, and as such it picks up more obvious commonality between a summary and a reference summary. BLANC is more 'semantic' and is closer to the experts. The same probably explains that BERTScore is on the same level as BLANC with the crowd sources, but is much worse with the experts. 

\section{Conclusion}
\label{sec:conclusion}

We illustrated how to effortlessly adapt the BLANC metric to a language other than English. We compared the BLANC scores to the human ratings and found the configuration giving the best correlations on the considered German dataset. We have shown that BLANC can perform on German as good as the most used automatic metric ROUGE when evaluating informativeness, and it performs comparable to the experts. However, our study has limitations because we conducted our analysis using only one monolingual human-annotated dataset. As part of future work, we will consider applying BLANC variations to more datasets, with a broader range of human annotations, and comparing them to more automated evaluation measures.

\bibliography{anthology,custom}
\bibliographystyle{acl_natbib}

\end{document}